\documentclass[pmlr]{jmlr}

\usepackage{longtable}
\usepackage{makecell}
\usepackage{graphicx}
\usepackage{siunitx}
\usepackage{float}
\usepackage{xcolor,color,soul,bm}
\usepackage{makecell}
\usepackage{booktabs}
\usepackage{amssymb}
\newcommand{\systemname}{UDAMA}
\newcommand{\imt}[1]{\textcolor{black}{#1}}
\usepackage{booktabs}
\usepackage{cleveref}
\usepackage{hyperref}
\crefname{subfigure}{}{}
 
\newcommand{\mlhc}[1]{\textcolor{black}{#1}}

\makeatletter
\def\set@curr@file#1{\def\@curr@file{#1}} 
\makeatother

\theorembodyfont{\upshape}
\theoremheaderfont{\scshape}
\theorempostheader{:}
\theoremsep{\newline}

\jmlrvolume{219}
\jmlryear{2023}
\jmlrworkshop{Machine Learning for Healthcare}

\title[UDAMA]{\textit{UDAMA}: Unsupervised Domain Adaptation through Multi-discriminator Adversarial Training with Noisy Labels Improves Cardio-fitness Prediction}

\author{\Name{Yu Wu\textsuperscript{1}} \Email{yw573@cam.ac.uk}\\
  \Name{Dimitris Spathis\textsuperscript{1,2}} \Email{ds806@cam.ac.uk}\\
  \Name{Hong Jia\textsuperscript{1}} \Email{hj359@cam.ac.uk}\\
  \Name{Ignacio Perez-Pozuelo\textsuperscript{3}} \Email{ip325@cam.ac.uk}\\
  \Name{Tomas I. Gonzales\textsuperscript{3}} \Email{tomas.gonzales@mrc-epid.cam.ac.uk}\\
  \Name{Soren Brage\textsuperscript{3}} \Email{soren.brage@mrc-epid.cam.ac.uk}\\
  \Name{Nicholas Wareham\textsuperscript{3}} \Email{nick.wareham@mrc-epid.cam.ac.uk}\\
  \Name{Cecilia Mascolo\textsuperscript{1}} \Email{cm542@cam.ac.uk}\\
  \addr \textsuperscript{1}Department of Computer Science and Technology, University of Cambridge, UK \\
  \addr \textsuperscript{2}Nokia Bell Labs, Cambridge, UK \\
  \addr \textsuperscript{3}MRC Epidemiology Unit, School of Clinical Medicine, University of Cambridge, UK
 }

\begin{document}

\maketitle

\begin{abstract}

Deep learning models have shown great promise in various healthcare \mlhc{monitoring} applications. 
\mlhc{However, most healthcare datasets with high-quality (gold-standard) labels are small-scale, as directly collecting ground truth is often costly and time-consuming. As a result, models developed and validated on small-scale datasets often suffer from overfitting and do not generalize well to unseen scenarios.}
At the same time, large amounts of imprecise (silver-standard) labeled data, 
annotated by approximate methods with the help of modern wearables and in the absence of ground truth validation, are starting to emerge. However, due to measurement differences, this data displays significant label distribution shifts, which motivates the use of domain adaptation. To this end, we introduce \textbf{\systemname{}}, a method with two key components: \textbf{Unsupervised Domain Adaptation} and \textbf{Multi-discriminator Adversarial Training}, where we pre-train on the silver-standard data and employ adversarial adaptation with the gold-standard data along with two domain discriminators.
\mlhc{In particular, we showcase the practical potential of \systemname{} by applying it to Cardio-respiratory fitness (CRF) prediction. CRF is a crucial determinant of metabolic disease and mortality, and it presents labels with various levels of noise (gold- and silver-standard), making it challenging to establish an accurate prediction model. Our results show promising performance by alleviating distribution shifts in various label shift settings. Additionally, by using data from two free-living cohort studies (Fenland and BBVS), we show that \systemname{} consistently outperforms up to 12\% 
compared to competitive transfer learning and state-of-the-art domain adaptation models, paving the way for leveraging noisy labeled data to improve fitness estimation at scale.}


\end{abstract}

\section{Introduction}
Deep learning (DL) has been widely applied to many healthcare applications, such as sleep stage classification~\citep{sleep-detection}, stress detection~\citep{stree-detection}, and fitness prediction such as cardiorespiratory fitness and exercise adherence prediction~\citep{fitness-prediction,zhou_2019}.
However, the significant breakthroughs and encouraging results brought by DL are often accompanied by the need for accurate data collection under laboratory-controlled experiments and clinically verified labeling, termed as \textit{gold-standard}. 

Collecting high-quality labels (i.e., gold-standard) for healthcare applications may require extensive effort and can be particularly time-consuming. For example, developing a precise epileptic seizure diagnosis model requires electroencephalography (EEG) during ambulatory screening~\citep{seizure} for accurate brain function detection and a diagnosis from physicians or neurologists to label seizure occurrence. Consequently, most existing datasets tend to be small-scale, leading to poor performance and model generalization~\citep{Raschka-2018} when developing DL models for different cohorts. 

\mlhc{In comparison, with the widespread use of mobile and wearable devices, less accurate yet large-scale labels, referred to as \textit{silver-standard}, are available. 
For example, Heart rate or Oxygen saturation (SpO$_2$) can be easily gathered from smartwatches in daily life without clinical visits. However, these silver-standard labels, which are derived from a less accurate estimation scheme, often lead to predictions with lower accuracy, as they are often noisy and display distribution shifts compared to gold-standard labels~\citep{karimi2020}.}


As such, \mlhc{gwojuedold-standard labels are crucial for the development and validation of robust clinical models. Although silver-standard labels with extensive labeling are easy to access, they usually contain noise due to less accurate collection schemes and are characterized by distribution shifts, which makes validation against gold-standard data difficult. In this paper, we answer the following question: \textit{Can we leverage large-scale noisy silver-standard datasets to improve deep learning model validation on gold-standard datasets for healthcare applications?}}

Domain adaptation (DA) is the natural candidate to solve the problem of distribution mismatch between the source (large-scale data) and target domains (small-scale data)~\citep{Patricia-2014}. Specifically, discrepancy-based and adversarial-based techniques are prevalent among recent DA methods and achieve state-of-the-art performance in image and language tasks~\citep{coral,wdgrl, Ganin-2015, Bousmalis-2016, Du-2020}. Discrepancy-based DAs seek to reduce the divergence between domains, especially the distance between two domains in the feature space, and demonstrate promising results~\citep{mmd,da_survey}. On the other hand, adversarial-based DAs employ domain discriminators to encourage domain confusion through an adversarial objective~\citep{da_survey, Ganin-2015}. \mlhc{However, these approaches mainly concentrate on large-scale and gold-standard source labels to adapt to unlabeled target data, while neglecting the fact that the source domain might contain noisy silver-standard labels.}

In this paper, we present \systemname{} and target the \textit{label distribution} shift problem from noisy labeled but large-scale source domains caused by less accurate data collection schemes to target domains with small-scale gold-standard labels. Our proposed framework is inspired by adversarial-based DA methods, which often contain a specific discriminator that categorizes the source or target domains from which samples originate. However, compared to existing approaches that mainly focus on discriminating domains as a binary classification task, \systemname{} captures the fine-grained information that resides within the label distribution shifts. To achieve this, we introduce a fine-grained discriminator that attempts to discriminate the distribution of domain labels, thereby capturing domain-invariant feature representation learning. As a result, through a multi-discriminator learning scheme, \systemname{} can effectively learn cross-domain representation by integrating both coarse-grained and fine-grained domain information, resulting in promising performance on small-scale datasets. 

\mlhc{We practically demonstrate the potential of \systemname{} in the context of healthcare data by applying it to cardiorespiratory fitness (CRF) prediction. CRF is a significant predictor of cardiovascular disease (CVD)~\citep{predictor}, one of the leading causes of death globally~\citep{Kaptoge2019}. CRF is directly measured by maximal oxygen consumption (VO$_2$max), which is assessed using heart rate responses to standard maximal exercises test (i.e., gold-standard). However, collecting VO$_{2}$max labels through such tests is difficult and expensive and needs extra equipment and clinical monitoring. Alternatively, submaximal VO$_{2}$max tests~\citep{gonzales2020submaximal} have been proposed to capture fitness levels. Despite their potential, such measurements have been shown to provide silver-standard labels with lower accuracy and a shift in distribution compared with gold-standard VO$_{2}$max.}


\mlhc{In this paper, we validate the power of silver-standard VO$_{2}$max to enhance the precision of fitness predictions by mitigating distribution shifts using the proposed \systemname{}. Specifically, we apply \systemname{} on a large CRF datasets~\citep{feland,bvs} and show that \systemname{} achieves the SOTA performance.} Our contributions are summarized as follows:
\begin{itemize}
    \item We propose a novel domain adaptation framework via multi-discriminator adversarial training, which incorporates samples from different distributions and allows us to learn better feature representations for (often small) gold-standard datasets.
    
    \item 
    \mlhc{In the CRF prediction task, gold-standard VO$_{2}$max are hard to get and silver-standard labels display domain shift problems. We are able to address the domain shift problem using \systemname{}, significantly improving the prediction accuracy in the target domain}. Furthermore, we stress-test our models with semi-synthetic data under various label shifts to show and test for robustness.
    
    \item 
    Through a set of extensive experiments, we show that \systemname{} achieves strong results (corr = 0.701 $\pm$ 0.032) and improves model performance up to 12.0\% compared to baselines on the CRF tasks. These results show that the proposed model is able to improve fitness prediction by only leveraging large-scale silver-standard labels. 
    
\end{itemize}

\subsection*{Generalizable Insights about Machine Learning in the Context of Healthcare}
\mlhc{In healthcare, high-quality datasets with gold-standard labels are expensive and hard to collect, leading to sparse and small-scale datasets that make it difficult to generalize to other unseen cohorts and applications. In contrast, collecting silver-standard labels from less accurate collection schemes with modern wearables is more affordable. However, these extensive less-accurate labels exhibit distribution discrepancies when compared to ground truth and cannot be directly leveraged for model deployment. Our work addresses this by introducing a multi-discriminator domain adaptation method for cross-domain representation learning, which transfers knowledge from large-scale weakly labeled data to small-scale health datasets with gold-standard labels. Specifically, in the context of CRF prediction, our results demonstrate that leveraging large-scale noisy VO${2}$max labels using \systemname{} not only achieves improved fitness prediction but also effectively mitigates label distribution shifts. This paves the way for the practical application of machine learning for real-world health outcomes. Furthermore, our approach can be easily adapted to various health-related tasks, particularly those involving high-dimensional time-series data and changes in label distribution for regression tasks.}

\section{Related Work}

\subsection{Cardio-fitness estimation}\label{2.1}
Numerous prediction models have been developed recently using different testing schemes (submaximal exercise / non-exercise tests) and a variety of machine learning methods~\citep{vo2max_ml,vo2max_ml2} to substitute direct measurements of VO$_{2}$max, \mlhc{the direct indicator of CRF}. Specifically, with the help of modern wearable technology, which can track physical activity, resting heart rate (RHR), and other biosignals, various silver-standard methods have emerged for more convenient VO$_{2}$max calculation without maximal exercise testing~\citep{wearble-vo2max,vo2max-non_2,vo2max-non_1,lack-vali-1,lack-vali-3}.
\mlhc{For example, deep learning models are utilized to predict fitness prediction converting raw wearable sensor data~\citep{Spathis-2022}. Nonetheless, these approaches overlook the impact of silver-standard labels generated from imprecise testing schemes, which can result in diminished model performance and untrustworthy fitness forecasts.}
Recently, a few works applying machine learning models on CRF predicting using gold-standard VO$_{2}$max labels have been proposed~\citep{Abut-2015}.
Nonetheless, they mainly test on small cohorts, which might lead to poor generalization performance. 
This paper proposes a novel DL method that aims to alleviate the distribution difference between imprecise silver-standard and gold-standard data for better fitness prediction (Results are discussed in \S\ref{6}).

\subsection{Domain adaptation}\label{2.3}
A domain combines the input population with the output following a certain probability distribution~\citep{Kouw-2018}. DA is one of the state-of-the-art solutions~\citep{Hoffman-2017} for learning information from an abundant labeled source domain and applying it to a new and unseen target domain with a different distribution. DA trains a feature extractor to learn the shared information across domains, so the model can generalize to new settings and thus mitigate the domain shift~\citep{Kouw-2018}. Discrepancy-based and adversarial-based DA approaches are effective among DA methods.

\subsubsection{Discrepancy-based method}
For discrepancy-based approaches, the network typically shares or reuses the initial layers between the source and target domains and aims to reduce feature space divergence~\citep{da_survey}. The maximum mean discrepancy (MMD)~\citep{mmd} is an effective distance metric that measures the distribution divergence between the mean embeddings of two distributions in the reproducing kernel Hilbert space (RKHS) to minimize the difference between two distributions. Many methods~\citep{mmd_1,mmd_2} utilize the MMD metric within the network to learn domain-invariant and discriminative representations. On the other hand, the correlation alignment (CORAL) ~\citep{coral} method is proposed to align the second-order statistics of the source and target distribution with a linear transformation. Deep-Coral~\citep{coral} is an extension of CORAL that can train a non-linear transformation to align the correlations of the representation embedding in deep neural networks from the source domain to the destination domain. Besides, some works minimize empirical Wasserstein distance between source and target feature representations and also showed good results in domain-invariant representation learning approaches~\citep{wdgrl}. Although these methods are effective and easily incorporated into deep neural networks, they are mainly designed for feature-based distribution shift alignment. Unlike most discrepancy-based methods, our method mainly addresses label distribution shift problems when the source domain has large-scale noisy labels.

\subsubsection{Adversarial-based DA}
Most adversarial-based methods are motivated by theory suggesting that a good cross-domain representation contains no discriminative information about the origin of the input and shows good performance to reduce domain discrepancy~\citep{Mei-2018}. Among these methods, Domain-Adversarial Neural Network (DANN)~\citep{Ganin-2015} first introduces adversarial training to domain adaptation. 
DANN utilizes a shared feature extractor to learn feature embedding and a discriminator to maximize the domain difference. After the convergence of the whole training, general features are learned for each input while their domain information cannot be discriminated. Followed by this, Domain Separation Networks (DSN)~\citep{Bousmalis-2016} and Adversarial Discriminative Domain Adaptation (ADDA)~\citep{Tzeng-2017} propose more sophisticated feature extraction methods using both shared and private feature extractors for the task classifier and domain discriminator. Moreover, Dual Adversarial Domain Adaptation (DADA)~\citep{Du-2020} uses two discriminators to pit against each other to aid the discriminative training.

The aforementioned methods mainly focus on discriminating domains as a coarse binary classification task while neglecting the rich information of the domain distribution. Moreover, the majority of applications of DA mainly focus on medical image segmentation and classification tasks ~\citep{Perone-2018, Mei-2020, Venkataramani-2018}. 
Our work aims to resolve the label distribution shift while \imt{keeping the input data constant for regression targets}.
We propose a novel DA framework (\systemname{}) with multi-discriminators to learn the coarse-grained and fine-grained domain information and validate it on a CRF prediction task.

\begin{figure}
    \centering
    \includegraphics[width=6 in]{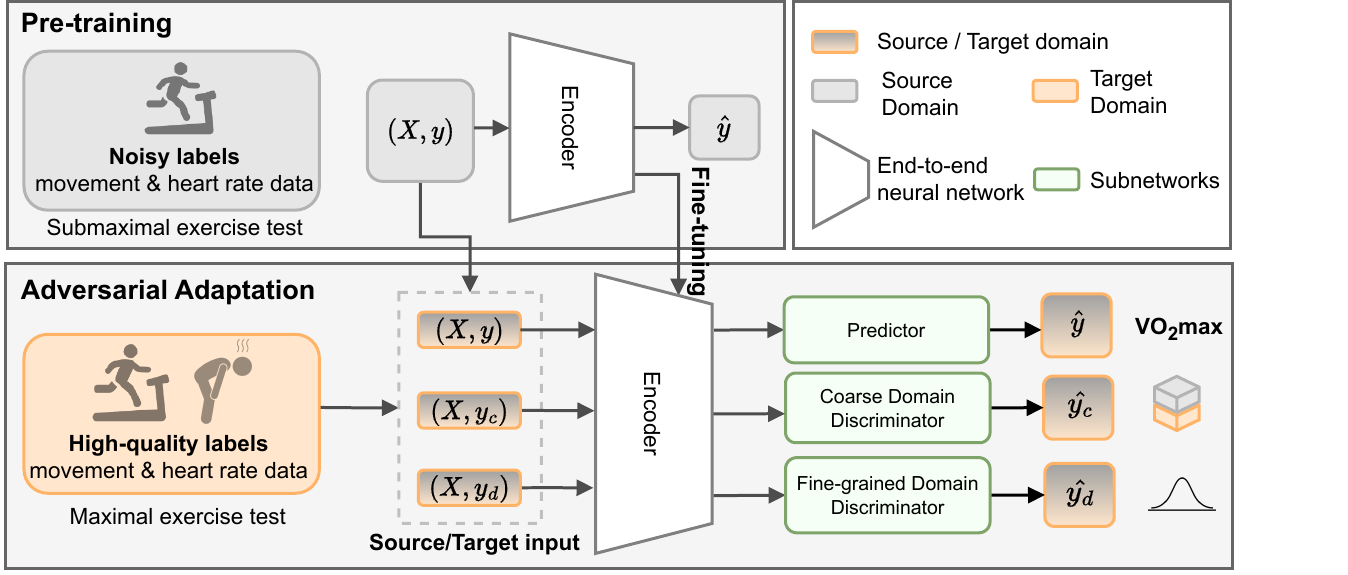}
    \caption{\imt{\textbf{ \systemname{} architecture and data pipeline.} }}
    \label{fig:workflow}
\end{figure}

\section{Cardio-respiratory Fitness Prediction}\label{4}
\mlhc{CRF is one of the strongest predictors of CVD compared with other risk factors like hypertension and type 2 diabetes~\citep{predictor}. Routinely assessing CRF through VO$_2$max, which is considered the benchmark measurement, provides valuable insights into a person's overall fitness. However, obtaining gold-standard VO$_2$max measurements, as shown in Figure~\ref{fig:workflow}, is time-consuming and thus rarely performed in clinical settings. In particular, it requires participants to undergo a maximal exercise test to reach exhaustion on a treadmill, while wearing a face mask with a computerized gas analysis system to monitor ventilation and expired gas fractions.}

\mlhc{Recently, less-accurate measurement schemes such as  sub-maximal exercise tests (silver-standard) utilizing modern wearables embedded with accelerometers and ECG sensors have started to provide opportunities for population-level fitness prediction. However, this alternative measurement method has been shown to demonstrate a measurement bias ranging from -3.0 to -1.6 ml O2/\textit{min}/\textit{kg} and a Pearson’s r ranging from 0.57 to 0.79~\citep{gonzales2020submaximal} compared to gold-standard. Apart from producing less accurate VO$_2$max values, these measurements also exhibit distribution mismatches, making it difficult to integrate into clinical practice.}

\mlhc{Similar to other healthcare applications, the distribution shift between silver- and gold-standard labels in the CRF prediction task is often ill-defined. 
To tackle this issue, this paper aims to adapt the source domain, characterized by noisy yet large-scale silver-standard labels, to the target domain with small-scale gold-standard datasets. Specifically, we introduce a novel adversarial-based unsupervised domain adaptation framework with multiple domain discriminators, i.e., \systemname{} to learn domain-invariant features and improve model validation on gold-standard VO$_2$max prediction. }

\section{Methods}
This work introduces a novel unsupervised domain adaptation framework that utilizes multi-discriminator during adversarial training. The overall model architecture and multi-discriminator training scheme are shown in Figure~\ref{fig:workflow}. 
Herein, in this section, we formally discuss the problem formulation (\S\ref{3.1}) and details of our framework, which includes the first-step pre-training and second-step multi-discriminator domain adaptation training (\S\ref{3.3}). 

\subsection{Problem Formulation and Notation}\label{3.1}
Here, we denote \(\bm{D_s}\) as the source domain containing silver-standard labels and \(\bm{D_t}\) as the target domain with gold-standard labels, as shown in Figure~\ref{fig:workflow}. 
For each domain, we assume the data as \bm{$X = (x_1, ..., x_n) \in \mathbb{R}^{N\times T \times F}$} corresponds to the accelerometer and Electrocardiogram (ECG) data from a chest ECG device \imt{
and a target regression VO$_2$max labels \bm{$y = (y_1, ..., y_n)\in \mathbb{R}^{N}$}. 
}
Additionally, we take into account contextual information such as the height or weight as metadata \(\bm{M = (m_1, ..., m_n)\in \mathbb{R}^{N \times F} }\). \imt{For the input data $\bm{X}$ and $\bm{M}$, \bm{$N$} represents the number of samples/subjects, \bm{$T$} represents the length of input sequences, and \bm{$F$} represents the number of input features.   }
Besides, \imt{
We use coarse- and fine-grained domain labels to train our model during adaptation and utilize multiple discriminators to differentiate between them. In particular, \bm{$y_c = (y_c[1], ..., (y_c[n])$} is the categorical value representing the coarse-grained binary domain label. \bm{$y_d = (y_d[1], ..., (y_d[n])$} is the numerical value that denotes the fine-grained domain distribution label. 
}
The coarse-grained domain discriminator is \(\bm{D_c}\) and the fine-grained domain discriminator is \(\bm{D_f}\). Also, for the training process, we denote the feature encoder with \(\bm{E}\) and the regression predictor with \(\bm{G_y}\). 
The overall networks thus can be represented as \(\bm{\hat{y_c}= D_c \cdot E }\), \(\bm{\hat{y_d} = D_f \cdot E }\) and \(\bm{\hat{y} = G_y \cdot E }\). The full table is shown in Table~\ref{tab:notation}.

\imt{\subsection{Unsupervised Domain Adaptation and Multi-discriminator Adversarial Training} \label{3.3}}
Domain adaptation is a method for learning a mapping between domains with distinct distributions, including data distribution shifts such as covariate shift, conditional shift, and label distribution ~\citep{da_survey}. In this paper, we propose \systemname{}, the unsupervised domain adversarial training, to address the label distribution shift problem,
particularly when the source domain contains numerous noisy labels. 

\imt{As shown in Figure~\ref{fig:workflow}, after pre-training on source domain with large-scale silver-standard labels, we first incorporate part of prior knowledge from the \(\bm{D_s}\) to create the adversarial training environment. Then we use the mixed silver-standard and gold-standard data to train the predictors and discriminator during the adaptation phase.}

\imt{
In particular, the adversarial training process consists of an encoder ($\bm{E}$), a VO2max label predictor($\bm{G_y}$), and two domain classifiers/discriminators ($\bm{D}$) designed for label shift problems by distinguishing both the domain and domain distribution information. During training, the fine- ($\bm{D_f}$)and coarse-grained ($\bm{D_c}$) discriminators are first optimized to identify the domain of each sample (i.e., max{$\bm{D_f,D_c}$}). In adversarial, the label predictor and encoder are then optimized to predict continuous fitness values from encoding(i.e., min{$\bm{E, G_y}$}).  The above-mentioned adversarial process will finally achieve the trade-off (i.e., the best prediction result in the most difficult-to-distinguish domain). 
}

\subsubsection{Coarse-grained discriminator}
The coarse-grained discriminator (\(\bm{D_c}\)) is similar to other DAs~\citep{Mathelin-2020,Zhao-2018} and follows the DANN~\citep{Ganin-2015} style. In other words, \(\bm{D_c}\) aims to discriminate the source of each data point \imt{in the mixture of pre-training and target labeled data} as a binary classification task, where 0 represents the data comes from the \(\bm{D_s}\) and 1 from \(\bm{D_t}\). In specific, After getting the representation matrix by fine-tuned feature extractor, two fully connected layers with the corresponding activation in \(\bm{D_c}\) are used to discriminate the rough binary domains labels and predict a probability vector \(\bm{\hat{y_c}}\). Let \(\bm{\hat{Y_c} = {\hat{y_c}[n]}}\) denote the predicted probability vectors for all the data points in (\(\bm{D_t}\)). The classification loss of the coarse-grained discriminator is defined as:
\begin{equation}
L_{CSE} = \sum_{N}l_c(y_c[n], \hat{{y_c}}[n])
\end{equation}
where \(l_c\) is the cross entropy loss of a single data point, and by optimizing \(\bm{L_{CSE}}\) for \(\bm{D_c}\), we can force the extractor to learn a general feature by maximizing such divergence.

\subsubsection{Fine-grained discriminator}

However, a simple binary classification task cannot properly represent the domain label distribution. Therefore, we augment adversarial training with a fine-grained discriminator (\(\bm{D_f}\)) to discriminate distribution differences. Specifically, instead of generating the binary domain labels (0 or 1) for the source or target domain for each sample, we construct a more complex pseudo-label (\(\bm{y_d}\)) to represent its domain label distribution. Based on our observation of the health outcome labels, which conform to the Gaussian distribution, as shown in Figure~\ref{distribution}, we then assign the (\(\bm{y_d}\)) for each training sample \imt{during adaptation}.
\imt{Specifically, $\bm{y_d}$ represents the mean and variance of the regression label distribution. This label is based on whether the sample is from the pre-training or target domain.
}
Therefore, after generating the feature matrix using \(\bm{E}\), \(\bm{D_f}\) is designed to distinguish the mean and variance of the label distribution using two fully connected layers with the corresponding activation. Let \(\bm{\hat{Y_d} = {\hat{y_d}}[n]}\) denote the predicted probability vectors for all the data points in (\(\bm{D_t}\)). Then, the loss of the fine-grained discriminator is defined as:
\begin{equation}
L_{GLL} = \sum_{N}l_g(y_d[n], \hat{{y_d}}[n])
\end{equation}
where \(l_g\) is each data point's Gaussian Negative Log-Likelihood (GLL) loss. In particular, GLL optimizes the mean and variance of a distribution and thus further maximizes the nuance changes among the sample and updates the discriminator.

After that, \(D_c\) and \(D_f\) can maximize the difference between the source and target domains for the multiple-domain discriminator training scheme. Meanwhile, the encoder and the predictor try to maximize the correct prediction of \(y\). These two modules play two games and finally reach a balance during the training. As a result, the encoder and predictor can learn a representation that cannot tell the difference between the source and target domains after the training is converged.

\subsubsection{Objective functions and training}
For adversarial training, the shared encoder first leverages the pre-trained model and extract general feature. Then discriminators are trained simultaneously to differentiate the domain labels. The predictor is used for the regression healthcare outcomes prediction task, and a mean squared error loss \(\bm{L_{MSE}}\) is applied to optimize the \(\bm{G_y}\). The detailed training overflow and input for each module are shown in Figure~\ref{fig:train_overflow}. Finally, the whole framework can be optimized by the total loss 
\(\bm{L}\), which is defined as:
\begin{equation}
L = \alpha L_{MSE} - \lambda_1L_{CSE} -\lambda_2L_{GLL}
\end{equation}
We optimize the overall loss \(\bm{L}\) to minimize the predictor loss while maximizing the loss of domain discriminators. In detail, $\alpha$ is used to scale down the predictor loss to the same level as predictors, $\lambda_1$ and $\lambda_2$ control the relative weight of the discriminator loss, and $\lambda_1$ + $\lambda_2$ = 1. Hyperparameter choice details are discussed in Section~\S\ref{experiment}.

\section{Experiments}\label{experiment}
We conduct various qualitative and quantitative evaluations to validate our models on the cardio-respiratory fitness prediction task with wearable sensing (\S\ref{5.1}). We outline the model's architecture and how it differs from other baselines (\S\ref{5.2} and \S\ref{5.3}). Next, we further quantify the impact of incorporating source domain knowledge to \systemname{} (\S\ref{5.4}).
The overall workflow is depicted in Figure~\ref{fig:workflow}. Code is available at: \url{https://github.com/yvonneywu/UDAMA}.

\subsection{Datasets and Training Strategy}\label{5.1}
\textbf{Datasets.} The source domain comes from the silver-standard measurement study (Fenland), including 11,059 participants. It contains a combined heart rate and movement signal from chest ECG sensor Actiheart and noisy VO$_{2}$max labels collected from submaximal exercise tests~\citep{feland}. 
The target domain \(\bm{D_t}\) represents the gold-standard measurement dataset BBVS, which is a subset of 181 participants from the Fenland study with directly measured gold-standard VO$_{2}$max~\citep{bvs} during the maximal exercise tests. In the BBVS study, participants need to wear a face mask to measure respiratory gas measurements~\citep{Rietjens-2001} to experience exhaustion tests.
Further, movement and heart rate are collected using the Actiheart (CamNtech, Papworth, UK) sensor. The details of the two datasets are in Appendix~(\ref{4})

\textbf{Training Strategy.} We evaluate \systemname{} on these two datasets by first pre-training a model on the \(\bm{D_s}\) Fenland. 
Second, we develop the adversarial training framework with multi-discriminators on the BBVS (\(\bm{D_t}\)), with the help of incorporated prior domain knowledge (i.e., injecting random samples from the source domain). For each domain, feature choices are listed in Appendix~\ref{apd:first}. 
After the adversarial adaptation, we predict VO$_{2}$max on the held-out test set of BBVS using 3-fold cross-validation using \systemname{}. Within each fold, the dataset is split into 70\% training and 30\% testing consisting only of target domain samples.

\subsection{Model architecture and tuning}\label{5.2}
\subsubsection{Model architecture} 
This section discusses the details of the neural networks used in this study. For the encoder network, two modules are integrated within the network to extract temporal and metadata information. In particular, we use two Bidirectional GRU layers of 32 units for the time series data module, followed by one 1D-global averaging pooling layer. On the other hand, in the meta-data module, MLP layers of dimensionality 128 are constructed to extract associated metadata representation after a Batch Normalization layer. Following this, the time and metadata module outputs are concatenated together to generate a complete embedding matrix for the subsequent regression or classification tasks. The training architecture is shown in Figure~\ref{fig:train_overflow}.

The training pipeline consists of two phases using the pre-training and fine-tuning learning scheme. First, after comparing different parameter-sharing techniques for fine-tuning the encoder utilized in the source domain, we freeze the first GRU and MLP layers and fine-tuned the remaining network. Compared with freezing all layers except re-training the output layer, the fine-tuning scheme in our network could capture the general features from the lower layers and extract problem-specific characteristics from higher layers. Lastly, the representation embedding produced from the fine-tuned encoder is transmitted to distinct tasks with a linear activation layer appropriate for predicting the fitness level or classifying the domain labels.

\subsubsection{Hyper-paremeter tuning} 
All network blocks in the framework are trained using Adam with a learning rate tuned over \{1e-2, 1e-3\}.
The dropout rate is tuned over the following ranges \{0.2, 0.3\}. Moreover, we tune the batch size between 8, 16, and 32 based on the efficiency and stability of the training process.
To tune the \systemname{} total loss, we conduct a grid search~\{0.01, 0.02, 0.03\} for the $\alpha$ and~\{(0.9, 0.1), (0.8, 0.2), (0.7, 0.3), (0.6, 0.4), (0.5, 0.5)\} for the combination of $\lambda_1$ and $\lambda_2$. We perform early stopping to combat overfitting until the validation loss stops improving after ten epochs. The details of hyperparameter selection are listed in Table~\ref{table:hyper}.

\subsection{Baselines}\label{5.3}
To verify the effectiveness of our proposed network, we compare the \systemname{} against several baselines:
\begin{itemize}
    \item \imt{\textbf{In-domain supervised model.} A multi-model network with training on the same domain train and test set. }
    \item \textbf{Out-of-domain supervised model}. A network with the same structure as the pre-training model, using wearable data and common biomarkers in \(\bm{D_t}\) as input to predict VO$_{2}$max. 
    \item \textbf{Transfer learning.}  A pre-trained model trained on \(\bm{D_s}\) is reused and fine-tuned on the target domain. 
    \imt{\item \textbf{Autoencoder~\citep{Srivastava_2015}.}  Pre-train a model with stacked recurrent autoencoders on \(\bm{D_s}\) and fine-tune the representation from the encoder to \(\bm{D_t}\). }
    \item \textbf{Deep-Coral~\citep{coral}}. A widely employed discrepancy-based domain adaptation minimizes the divergence between the source and target in feature space. In particular, it seeks to align the second-order statistics of the source and target distributions.
    \item \textbf{WDGRL~\citep{wdgrl}}. Wasserstein Distance is used to minimize the disparity between the source and target representations in the feature space. Specifically, the distance will be optimized in an adversarial way for the feature extractor, and domain-invariant features will be learned.
    \item \textbf{Domain Adversarial Neural Networks (DANN)~\citep{Ganin-2015}.} A benchmark domain adaption method that uses adversarial training for binary domain classification. 

    
\end{itemize}
\mlhc{Although most recent domain adaptation methods have shown enhanced performance in dealing with distribution shift through self-training or sophisticated adversarial training scheme~\citep{Du-2020, Liu_2021}, they do not specifically tackle the regression tasks or healthcare datasets that contain both noisy and gold-standard labels. Therefore, their relevance in the cardio-fitness prediction task is limited. }

\subsection{Effect of Injected Source Domain Samples}\label{5.4}
Our proposed approach incorporates prior knowledge from the source domain to create the adversarial training environment. To assess the degree to which the incorporated source domain knowledge impacts our model, we evaluate \systemname{} on $\bm{D_t}$ with different levels of 
injected samples from ($\bm{D_s}$). As such, we put \{0.1\%, 0.2\%, 0.4\%, 1\%, 2\%, 4\% \} from $\bm{D_s}$, which equals to \{1\%, 5\%, 10\%, 30\%, 50\% and 100\%\} of training data of the target domain ($\bm{D_t}$). We set the maximum amount to 100\% because the two domains are highly different, and for any larger amount, the model would be trying to predict the silver data rather than the gold data. Specifically, the number of samples differs between source and target domain samples (source/target = 25:1), which means 0.2\% of source data equals 5\% of target data. If we continually add more source data, for example, 4\% of the source (which equals 100\% of the target data), we will ultimately train on the source data instead of the target data.
Finally, we run the experiments 15 times with different seeds to assess the impact of the injected noisy samples using the average performance on the VO$_{2}$max prediction task. Results are shown in \S\ref{6.3}.

\subsection{Metrics}
The prediction performance of all models is evaluated based on standard regression evaluation metrics such as the Mean Squared Error (MSE) and Mean Absolute Error (MAE). The coefficient of determination (R$^2$) and the Pearson correlation coefficient (Corr) are also used to evaluate the model performance on the health-related outcome prediction task.

\section{Results and Discussion}\label{6}
In this section, we present the results of applying \systemname{} to the cardio-fitness prediction task and compare it with the baselines (\S\ref{6.1}). Additionally, we discuss the performance in addressing the domain shift problem (\S\ref{6.2}) and the impact of injected source domain knowledge (\S\ref{6.3}). Furthermore, we conduct an ablation study to verify the effectiveness of our framework's structure (\S\ref{6.4}). Finally, we discuss the model robustness in \S\ref{6.5} with semi-synthetic data.

\begin{table}
  \caption{\textbf{Evaluation of different methods on CRF prediction task.} Each result displays the mean value with standard deviation from three-fold cross-validation. In particular, \textbf{In-domain} means the model is trained on the same domain, namely trained on \(D_t\) and tested on \(D_t\), while \textbf{Out-of-domain} corresponds to models trained on Fenland (\(D_s\)) and  adapted/fine-tuned to BBVS (\(D_t\)). All \textbf{Out-of-domain} models are evaluated on the BBVS test set.}
  \resizebox{1\textwidth}{!}{%
  \begin{tabular}
  {llllll}
  \toprule
    \textbf{In-domain} &\textbf{Training method} & \textbf{R$^2$} & \textbf{Corr} &  \textbf{MSE} & \textbf{MAE} \\ 
    \midrule
    \(D_t \rightarrow  D_t\) & \textit{Supervised} & \(0.123 \pm 0.111\)  & \(0.622 \pm 0.036\) & \(43.778 \pm 6.012\)   & \(5.263 \pm 0.277\)  \\
    \toprule
    \textbf{Out-of-domain} &\textbf{Training method} & \textbf{R$^2$} & \textbf{Corr} &  \textbf{MSE} & \textbf{MAE} \\
    \midrule
     \(D_s \rightarrow  D_t\) & \textit{Supervised} & \(-0.096 \pm 0.100\) & \(0.007 \pm 0.250\) & \(58.048 \pm 10.061\)  & \(6.336 \pm 0.621\)  \\ 
     &WDGRL ~\citep{wdgrl} & \(-0.100   \pm 0.073\) & \(0.004 \pm 0.161\) & \(55.611 \pm 10.61\) & \(6.044 \pm 0.615\) \\
     & Autoencoder~\citep{Srivastava_2015} & \(-0.067 \pm 0.069\) & \(0.127 \pm 0.222\) & \(53.254 \pm 4.878\) & \(5.973 \pm 0.194\) \\
    &Deep-Coral~\citep{coral} & \(0.021 \pm 0.073\)  & \(0.360  \pm 0.057\) & \(49.044 \pm 6.553\) &  \(5.638 \pm 0.374\)\\
    &Transfer learning (TF)  & \(0.283 \pm 0.037\)  & \(0.621 \pm 0.012\) & \(35.399 \pm  5.910\) & \(4.744 \pm 0.433\) \\ 
  &DANN~\citep{Ganin-2015} & \(0.288 \pm 0.077\)   & \(0.617 \pm 0.037\) & \(35.458 \pm 3.920\) & \(4.679 \pm 0.382\)  \\
  \hline
    &\textbf{UDAMA (ours)} & \textbf{0.459} \(\pm\) \textbf{0.063} & \textbf{0.701} \(\pm\) \textbf{0.032} &  \textbf{27.469} \(\pm\) \textbf{6.456} & \textbf{4.111} \(\pm\) \textbf{0.353} \\
    
    \bottomrule
  \end{tabular}%
}
  \label{table-results}
\end{table}

\subsection{Fitness prediction}\label{6.1}
We took 60 participants from the BBVS dataset as test samples to predict their CRF by predicting VO$_{2}$max values. The comparison between the proposed domain adaptation framework and baseline approaches is shown in Table~\ref{table-results}.

\imt{First, in the out-of-domain comparison, adversarial-based DA or transfer learning shows better performance than the discrepancy-based method under label distribution shift.} In particular, the discrepancy-based method, Deep-coral, increases the Corr and MAE to 0.36 and 5.638, respectively. Meanwhile, WDGRL aims to minimize the feature difference by employing Wasserstein distance, yielding results comparable to the out-of-domain supervised method, which directly applies the model trained on Fenland for testing BBVS. In contrast, the adversarial-based methods here display better results. DANN learns a representation that is predictive of the regression task but uninformative to the input domain and improves the Corr and MAE to 0.617 and 4.679, respectively. According to Corr and MAE, methods such as transfer learning with fine-tuning techniques also improve performance, achieving 0.621 and 4.744, compared to the discrepancy-based method. 

\imt{In general, high Corr and R2 values and low MSE and MAE demonstrate the model's ability to leverage noisy, large-scale labeled VO2max data for gold-standard VO2max prediction under label shift. However, the limited size of the test set might result in increased uncertainty during model evaluation. The results from both TF and DANN are similar as shown in Table~\ref{table-results}}. In contrast, \systemname{} utilizing both fine-tuning and adversarial-based domain adaptation methods outperforms all the abovementioned baselines. We observe that the correlation (Corr) outperforms the basic transfer learning methods by 12.9\%, the MSE increases by 22.4\%, and R$^2$ improves by 62.2\%. Moreover, compared with the in-domain supervised training on $\bm{D_t}$, \systemname{} shows a significant increase, improving R$^2$ from 0.123 to 0.459. 
Our method also achieves good performance when generalizing the model from the in-domain to out-of-domain BBVS setting, compared with the dramatic performance drop-down, as shown in Table~\ref{table-results}. Therefore, \systemname{} can leverage the large-scale noisy datasets information and alleviate the model performance degeneration performance compared with directly validating models on small-scale sensing datasets. 
\begin{figure}{
\centering  
\subfigure{
    \includegraphics[width=2.5 in]{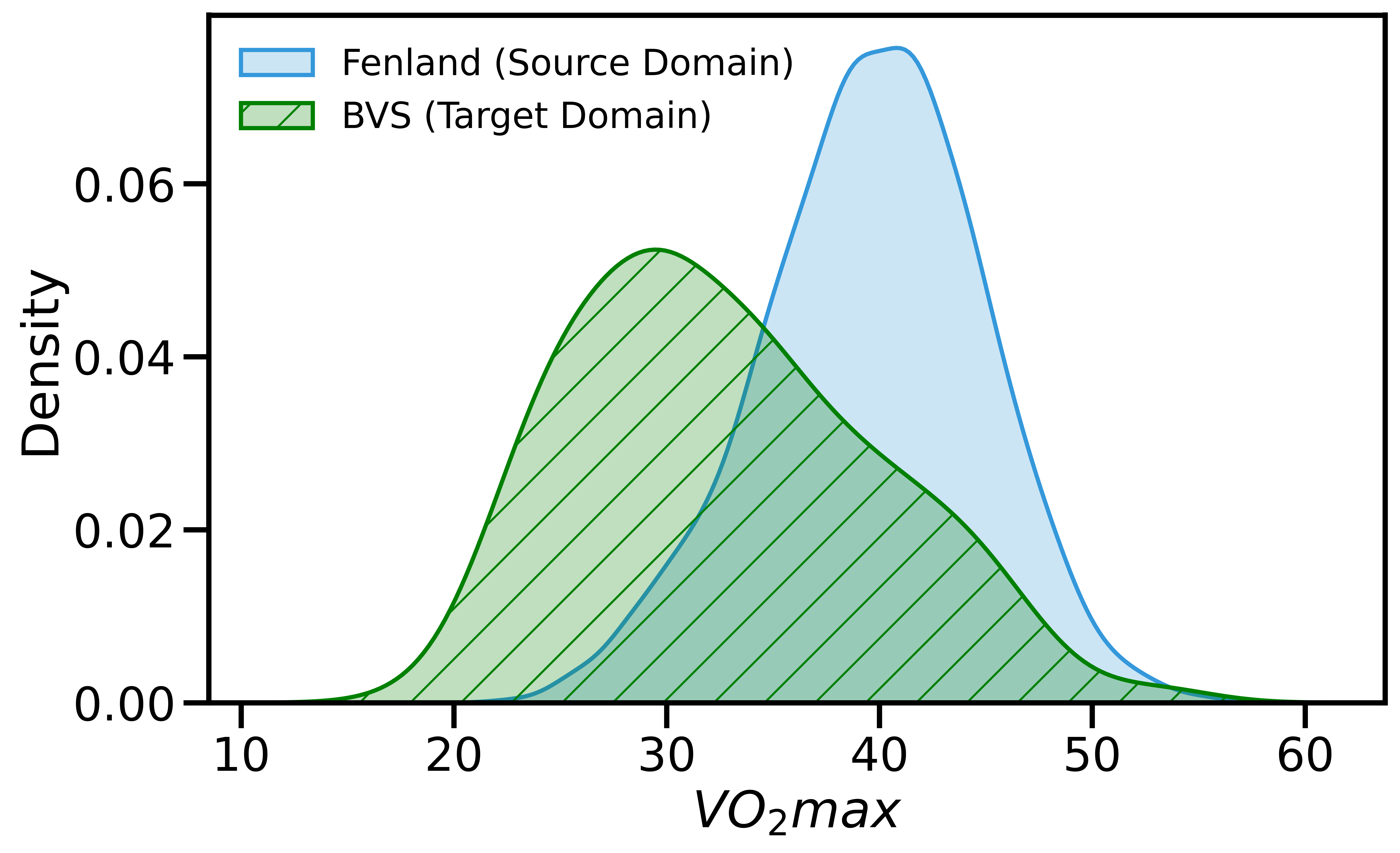}
    \label{fig:ori-distribution} 
    }
\centering
\subfigure{
    \includegraphics[width=2.5 in]{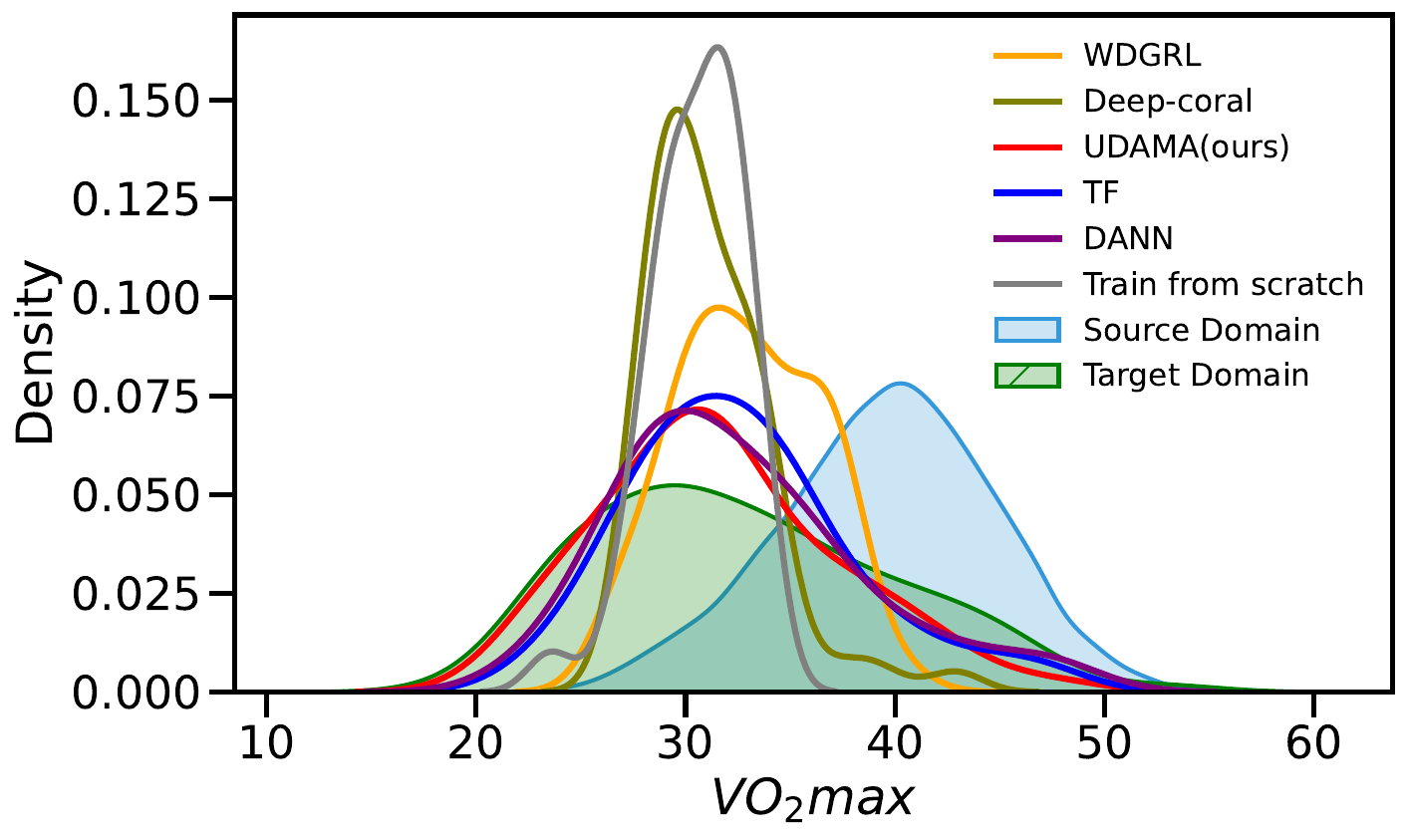} 
    \label{fig:domain_shift} 
    }
    \caption{\textbf{The distribution of BBVS, Fenland, and distribution of prediction of BBVS test set using different methods}. \textbf{(a)} Left figure shows the distribution of BBVS and Fenland datasets. \textbf{(b)} Right figure shows the prediction distribution of the BBVS test set from different methods.
    }
    \label{distribution}}
\end{figure}

\subsection{Domain shift}\label{6.2}
\begin{figure}{
\centering  
\subfigure{
    \includegraphics[width=2.5 in]{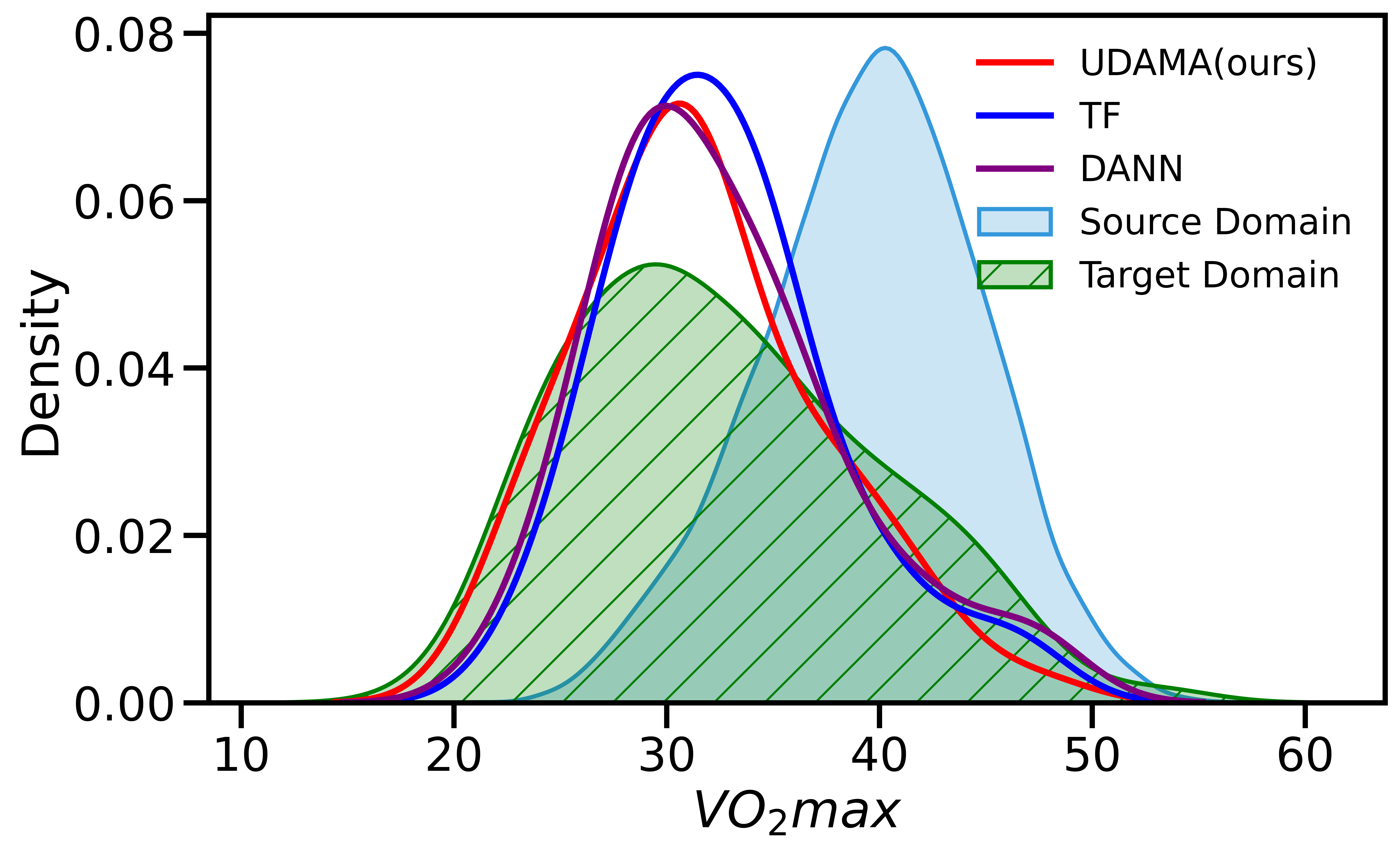}
    \label{fig:distribution-shift_good} 
    }
\centering
\subfigure{
    \includegraphics[width=2.5 in]{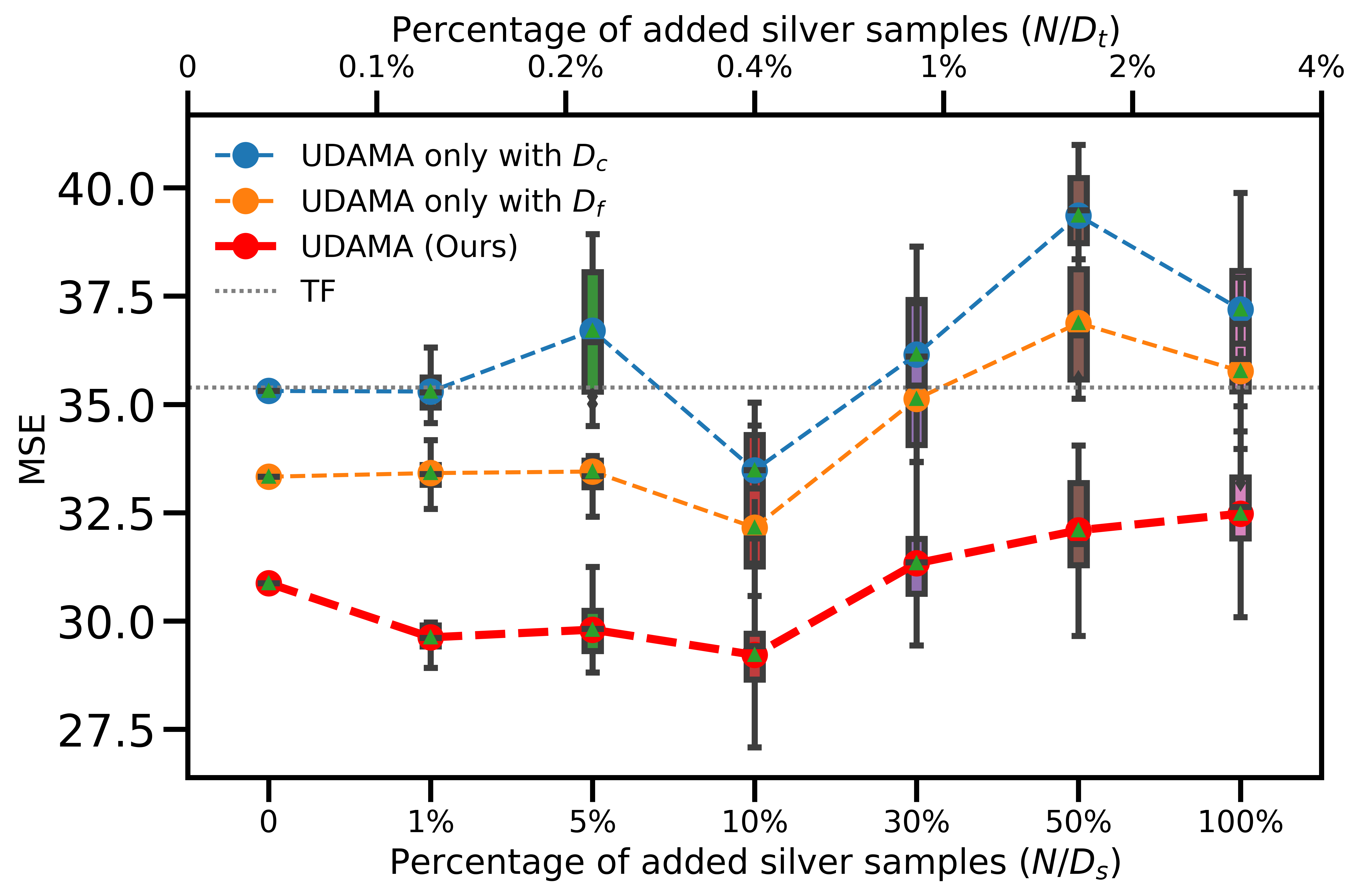}
    \label{fig:scale} 
    }
    \caption{ \textbf{(a)} Left figure shows \systemname{} mitigates distribution shifts.\textbf{(b)} Right figure shows the impact of injected source domain samples.
    }
    }
\end{figure}

To better compare whether \systemname{} or baseline DA methods can solve the label distribution shift problem effectively. Figure~\ref{distribution} presents the predicted label distribution of the BBVS test set from different methods. 
First, it shows that the \(\bm{D_s}\) dataset (i.e., Fenland) shares a different VO$_{2}$max underlying distribution compared to \(\bm{D_t}\) BBVS. Our results demonstrate that \systemname{} can learn the small dataset distribution during the adaptation phase and achieves promising results compared with other methods. Besides, we observe that both adversarial-based methods and transfer learning capture the mean and range of target domain distribution as shown in Figure~\cref{fig:distribution-shift_good}, whereas the discrepancy-based methods fail to learn the general distribution. We attribute this performance degeneration to the fact that discrepancy-based approaches, mainly designed to minimize the divergence between feature spaces, cannot alleviate the impact of noisy labeling.

Moreover, we use the Hellinger Distance (HD), which calculates the similarity of distributions between prediction and ground truth to examine the distance between two label distributions. Specifically, our framework's prediction of fitness level lies in the same range as the ground truth, while methods like Deep-coral or WDGRL fail at learning within this range. Besides, the distribution of \systemname{} ties close compared to baseline methods, where the normalized HD for \systemname{} is 0.179, and HD for TF is 0.264, for Deep-coral is 0.305. These results indicate that our framework effectively alleviates the distribution shift problem of the VO$_{2}$max prediction task and \systemname{} can leverage noisy silver-standard data to improve the performance on the gold-standard dataset. 

\subsection{Impact of Injected Knowledge from Source Domain}\label{6.3}
Instead of using a generator, we incorporated source domain knowledge to create the adversarial training environment. Figure~\cref{fig:scale} shows the average performance of adding the different scales of injected source domain samples to the target domain.
Each box plot shows the average MSE results of 15 runs with added random samples from the source domain. 
As shown in Figure~\cref{fig:scale}, our method performs better than the baseline methods with added samples in all cases, even after adding 100\% of noisy samples of $\bm{D_t}$ from $\bm{D_s}$.
Specifically, we observe that it achieves the best results and showcases the most positive transfer by only adding 0.4\% of the source domain, which equals 10\% of \(\bm{D_t}\). In contrast, if we continue to add more noisy information to the adaptation stage, the performance will gradually decrease as the source and target domain data reach the ratio of 1:1. After that, the adaptation tends to learn the noisy source domain representation instead of the target domain, yielding a negative transfer. Therefore, only injecting a few samples from the source domain to create the adversarial training environment might help to learn more domain-invariant features and achieve optimal results.


\subsection{Ablation Study} \label{6.4}
Our framework comprises two joint discriminators, so we perform an ablation study to understand each discriminator's effect. Based on the observation that incorporating a 10\% amount of BBVS from the \(\bm{D_t}\) Fenland dataset achieves the best transfer, we train different discriminators under this setting. We observe that a single discriminator (coarse-grained or fine-grained) exhibits more competitive performance than the baseline TF or DANN methods, as shown in Table~\ref{table:ablation study} and Figure~\cref{fig:scale}. Specifically, the fine-grained discriminator alone significantly outperforms the baseline results: +15.8\% TF (based on MSE). This indicates that utilizing a fine-grained distribution domain discriminator, which discriminates the domain label distribution, enhances the training framework. Therefore, with the combination of \(\bm{D_f}\) and \(\bm{D_c}\), \systemname{} can learn the cross-domain information without capturing the domain information of the input and alleviate the noisy labeling problem. 

\imt{Additionally, our method is unique in that it uses wearable-based CRF prediction task, unlike other methods that solely rely on anthropometric data~\citep{Nes_2011}. Our experiments have demonstrated that combining sensor data with anthropometric information leads to improved prediction accuracy. Although the sensor data alone is unreliable and insufficient for accurate VO${2}$max prediction, when combined with anthropometric data, the accuracy of \systemname{} increases from 0.679 (using only anthropometric data) to 0.701 (using anthropometric data and wearable sensor data such as acceleration, heart rate, and heart rate variability). As a result, using anthropometric and low-cost wearable devices together enables more accurate VO$_{2}$max prediction through \systemname{}.  }

\begin{table*}
  \centering 
  \small 
  \caption{\textbf{Ablation study by removing one of the discriminators.
  }}
  \begin{tabular*}{\textwidth}{l @{\extracolsep{\fill}} llll}
  \toprule
    \textbf{Discriminator} & \textbf{R$^2$} & \textbf{Corr} & \textbf{MSE} & \textbf{MAE} \\
    \midrule
    Coarse-grained & 0.339 $\pm$ 0.053 & 0.626 $\pm$ 0.032 & 33.485 $\pm$ 6.740  & 4.580 $\pm$ 0.337 \\ 
    Fine-grained  & 0.394 $\pm$ 0.030 & 0.666 $\pm$ 0.030 & 30.578 $\pm$ 5.300  & 4.498 $\pm$ 0.324 \\ 
    \textbf{UDAMA (ours)} & \textbf{0.459} $\pm$ \textbf{0.063} & \textbf{0.701} $\pm$ \textbf{0.032} &  \textbf{27.469} $\pm$ \textbf{6.456} & \textbf{4.111} $\pm$ \textbf{0.353} \\
    \bottomrule
  \end{tabular*}

  \label{table:ablation study} 
\end{table*}


\begin{table*}
  \centering 
  \small 
  \caption{\imt{\textbf{Simulated Label Distribution Shift}. Kullback–Leibler divergence (KL divergence) calculates the distribution difference between the shifted $D_s$ and fixed $D_t$.}
  }
  \begin{tabular*}{\textwidth}{l @{\extracolsep{\fill}} llll}
  \toprule
      Source dataset& \textbf{KL divergence} & \textbf{UDAMA Corr} & \textbf{DANN Corr }  \\
    \midrule
    Fenland  & 0.461  & 0.701 $\pm$ 0.032  &  0.617 $\pm$ 0.037 \\ 
     Left shifted Fenland & 1.607 & 0.656 $\pm$ 0.065 &  0.589 $\pm$ 0.027 \\ 
  Right shifted Fenland & 3.188  & 0.646 $\pm$ 0.035  &  0.608 $\pm$ 0.037  \\
    \bottomrule
  \end{tabular*}

  \label{table:simulation} 
  \vspace{-0.1in}
\end{table*}

\subsection{Robustness assessment with semi-synthetic data shift}\label{6.5}
\mlhc{Although the distribution shift among gold- and silver-standard labels are prevalent in healthcare applications, there are very few open available datasets with such a challenge where we can apply our method. Therefore, to further evaluate the efficacy of \systemname{}, we generate semi-synthetic datasets with various label distribution shifts for the cardio-fitness prediction.}

Motivated by the label distribution shift simulation for classification, as seen in ~\citep{Lipton_2018}, 
we shift the labels in the $\bm{D_s}$ by a fixed offset and Gaussian noise. By conducting experiments with varying degrees of label shifts, we gain a comprehensive understanding of the effect of label shifts on CRF prediction tasks. As the shift becomes greater from the target domain, the performance decreases. In particular, we show two extreme cases by pushing the source domain shift to left and right to stress-test \systemname{}, and the results are shown in Table ~\ref{table:simulation}. Despite the increasing KL divergence, which indicates a greater deviation from the ground truth data, our method still displays robust performance compared to the baseline DA method, especially in the case of low fitness on the left side. These stress tests highlight the versatility and robustness of our model in dealing with different input distributions.

\section{Discussion}
While deep learning has strong potential in numerous healthcare applications, the quality and size of labeled datasets remain a significant problem due to the high cost and labor-intensive nature of data collection. Consequently, the majority of high-quality datasets with gold-standard labels are small-scale, making it challenging to train efficient and generalizable models. Using large-scale wearable sensing datasets with inaccurate labeling could enhance the performance of such tiny datasets' models. However, distribution shifts, such as cohort and task shifts, may also contribute to performance degradation during model generalization and validation.
In this work, we proposed the unlabeled domain adaptation via multi-discriminator adversarial training framework (\systemname{}) to address the problem by leveraging a large-scale noisy silver-standard dataset. Our proposed method alleviates the domain shift problem and improves the performance of the challenging CRF prediction task. 
Moreover, our method has substantial potential to be applied to various other healthcare scenarios that demonstrate similar characteristics. While we have not specifically explored these applications due to limited benchmark datasets, the robust performance of our method under various distribution shifts using diverse datasets, containing both real and semi-synthetic data between gold-standard and noisy labels, shows the proposed method has the ability to generalize. We believe that the proposed method is applicable to a multitude of healthcare domains such as sleep monitoring, continuous monitoring of chronic diseases, and mental health research

\acks{This work was supported by ERC Project 833296 (EAR) and by Nokia Bell Labs.}

\bibliography{main}

\newpage
\appendix
\section{Notation Table and Training Structure}
We provide the details of the notation table and training structure here. 
\begin{table}[H]
  \centering 
  \caption{\textbf{Notation.}}
  \begin{tabular}{ll}
  \toprule
    \textbf{Notation} & \textbf{Description} \\
    \midrule
    \(\bm{D_s}\) & source domain with noisy (silver-standard) labels \\ 
    \(\bm{D_t}\) & target domain with high-quality (gold-standard) labels \\ 
    \(\bm{X \in \mathbb{R}^{N\times T \times F} }\) & input time-series sequences\\
    \(\bm{M \in \mathbb{R}^{N \times F} }\) & input user metadata \\
    \(\bm{N}\) & number of samples \\
    \(\bm{T}\) & length of input sequences \\
    \(\bm{F}\) & number of features \\
    \(\bm{y_c \in \mathbb{R}^{N}}\) & categorical value representing the coarse-grained binary domain label \\
    \(\bm{y_d \in \mathbb{R}^{N}}\) & numerical value that denotes the fine-grained domain distribution label \\
    \(\bm{y \in \mathbb{R}^{N}}\) & numerical scalar for task target value \\
    \(\bm{D_c}\) & coarse-grained domain discriminator \\
    \(\bm{D_f}\) & fine-grained domain discriminator \\
    \(\bm{G_y}\) & regression predictor \\
    \(\bm{E}\) & feature encoder \\
    
    \bottomrule
  \end{tabular}
  \label{tab:notation} 
\end{table}

\begin{figure*}

    \centering
    \includegraphics[width=6 in]{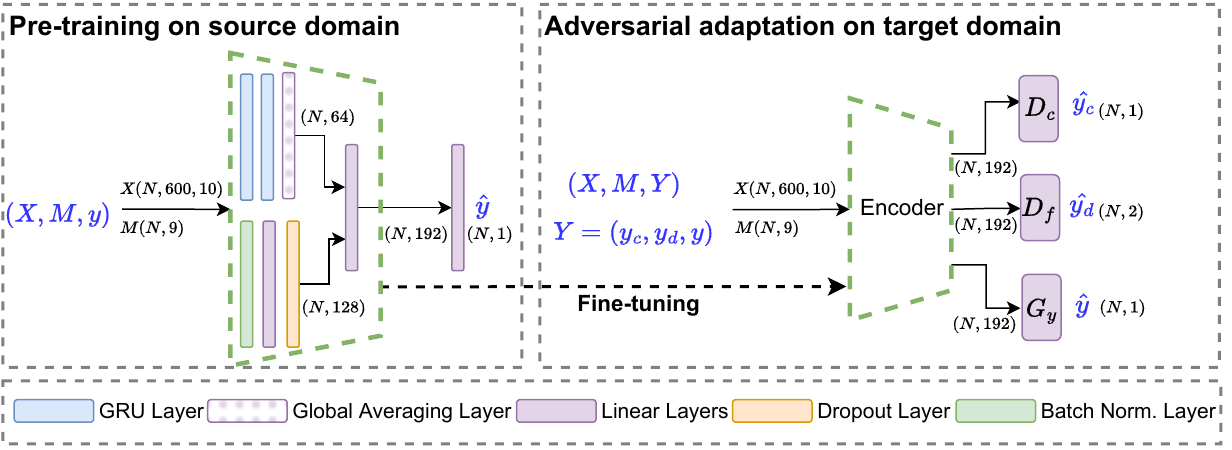}
    \caption{\textbf{\systemname{} detailed model architecture.}
    Visualization of the detailed network structure and the input and output dimensionality for each module. 
    }
    \label{fig:train_overflow}

\end{figure*}

\newpage
\section{Evaluation} 
First, we describe the cohort selection of the VO$_{2}$max prediction task (\S\ref{4.1}). Then, we discuss the \imt{data extraction and different measurement VO$_{2}$max} (\S\ref{4.2}) and feature extraction (\S\ref{4.3}) stage that occurs before the construction of the deep neural network. 

\subsection{Cohort Selection} \label{4.1}
Two datasets, Fenland and BVS, are utilized for experiments and evaluations in our work. Table~\ref{tab:cohortcharacteristics} indicates the descriptive characteristics of the two analysis samples. In addition, the mean and standard deviations for each characteristic are presented in this table.

\begin{table}[H]
\centering
\setlength\tabcolsep{2pt}

\caption{\textbf{Characteristics for study analytical samples: Fenland and BBVS study.}}
\begin{tabular}{lllllllllllll} 
\cmidrule[\heavyrulewidth]{1-12}
                                        & \multicolumn{5}{c}{\textbf{Fenland Study }}                                  &  & \multicolumn{5}{c}{\textbf{BBVS }}                                      &   \\ 
\cline{2-6}\cline{8-12}
                                        & \multicolumn{2}{r}{Men (n= 5229)}   &  & \multicolumn{2}{r}{Women (n= 5830)} &  & \multicolumn{2}{r}{Men (n= 98)} &  & \multicolumn{2}{r}{Women (n= 83)} &   \\ 
\cline{2-3}\cline{5-6}\cline{8-9}\cline{11-12}
                                        & \textit{mean}        & \textit{std} &  & \textit{mean} & \textit{std}        &  & \textit{mean} & \textit{std}    &  & \textit{mean} & \textit{std}      &   \\ 
\cline{1-12}
\textbf{Demographics}                   &                      &              &  &               &                     &  &               &                 &  &               &                   &   \\
Age (years)                              & 47.70 & 7.57 && 47.66 & 7.36               &  & 53.59         & 7.31            &  & 54.39         & 6.63              &   \\
\textbf{Anthropometrics}                & \multicolumn{1}{r}{} &              &  &               &                     &  &               &                 &  &               &                   &   \\
Height (m)~ ~                           & 1.78                 & 0.07         &  & 1.64          & 0.06                &  & 1.79          & 0.07            &  & 1.64          & 0.06              &   \\
Body mass (kg)                          & 85.85 & 13.83 && 70.54 & 13.92               &  & 84.63         & 10.15           &  & 69.31         & 10.95             &   \\
BMI (kg/m2)                             & 27.16 & 3.97 && 26.17 & 4.97                &  & 26.54 & 2.94            &  & 25.72         & 3.44              &   \\
\textbf{Physical activity}              &                      &              &  &               &                     &  &               &                 &  &               &                   &   \\
MVPA (min/day)                           & 35.87 & 22.35 && 34.40 & 22.59               &  & 40.97         & 25.23           &  & 41.73         & 22.26             &   \\
VPA (min/day)                           & 3.27 & 8.57 && 3.31 & 15.67                &  & 5.94          & 12.61           &  & 4.21          & 8.76              &   \\
\textbf{Resting Heart Rate}             &                      &              &  &               &                     &  &               &                 &  &               &                   &   \\
RHR (bpm)                              &61.48 & 8.68 & & 64.46 & 8.28               &  & 59.60        & 8.06            &  & 61.91         & 6.93              &   \\
\textbf{Cardiorespiratory fitness}      &                      &              &  &               &                     &  &               &                 &  &               &                   &   \\
VO2max (ml O2/\textit{min}/\textit{kg}) & 41.95 & 4.61 && 37.44 & 4.73                &  & 35.69         & 6.99            &  & 29.60         & 5.80              &   \\
\cmidrule[\heavyrulewidth]{1-12}
\end{tabular}

\label{tab:cohortcharacteristics}
\end{table}

\subsubsection{Fenland Study} The Fenland dataset with large-scale weakly-labeled VO$_{2}$max is used as the source domain (\(D_s\)) in this work. In particular, Fenland study~\citep{feland} is a prospective population-based cohort study of individuals aged 35-65 investigating the interaction between environmental and genetic factors in determining obesity, type 2 diabetes, and related metabolic disorders. The study has collected from 12,435 participants around Cambridgeshire in the East of England between 2005 and 2015. After a baseline clinic visit, participants were instructed to wear the chest sensor Actiheart for six consecutive days to collect the heart rate and movement data. Data from 11,059 participants were included in this study following the exclusion of those with insufficient or corrupt data or missing variables based on \S\ref{4.3}.


\subsubsection{Biobank Validation Study (BBVS)} BBVS dataset is the target domain (\(D_t\)) with clean gold-standard VO$_{2}$max labels~\citep{bvs}. The BBVS dataset is a subset of 191 participants from the Fenland study and aims to validate CRF measurement methods for the UKB-CRF test. During the study, similar electrocardiography~(Cardiosoft) and an Actiwave CARDIO device~(CamNtech, Papworth, UK) were utilized to collect signals. In particular, participants need to wear face masks with a computerized metabolic system to measure the VO$_{2}$max during the study. Similarly, data from 181 participants were included after excluding participants with insufficient data. All participants from two datasets provided written informed consent, and the University of Cambridge Ethics Committee approved the study.

\subsection{Data extraction and Cardiorespiratory fitness assessment} \label{4.2}
\imt{For both Fenland and BBVS study, participants wore the two standard Electrocardiography (ECG) electrodes Actiheart devices attached to the chest, which measured heart rate and movement recording at 60-second intervals~\citep{actiheart} for the wearable free-living sensing. }
The monitoring phase lasted for a total of six days.

In the Fenland study, VO$_{2}$max was measured using a previously
validated submaximal treadmill test\citep{gonzales2020submaximal} and considered as weak (silver-standard) labeling~\footnote{ https://www.mrc-epid.cam.ac.uk/research/studies/fenland/fenland-study-phase-1/}. 
\imt{During a lab visit, all participants conducted a treadmill test to determine their heart rate response to a submaximal test, informing their VO$_{2}$max by using a linear regression method.} 
HR was recorded for 15 minutes during the exercise, and RHR was calculated as the mean heart rate measured during the last 3 minutes. The RHR is a mix of the RHR mentioned above and the Sleeping HR as recorded by the ECG during the free-living period. \imt{Although using the submaximal test and linear regression capture the fitness level, measurement bias from -3.0 to -1.6 ml O2/\textit{min}/\textit{kg} and a Pearson’s r ranging from 0.57 to 0.79.  }

In comparison, VO$_{2}$max of participants in the BBVS dataset were directly measured during a maximal exercise test~\footnote{https://www.mrc-epid.cam.ac.uk/research/research-areas/physical-activity/}which was completed to exhaustion~\citep{bvs}, . Following standardized techniques used for the UKB-CRF test~\citep{CRF-manual}, participants did 5 maximal exercise tests to elicit VO$_{2}$max, including a UKB flat test, two UKB ramped tests with varying ramp rates and a steady-state test (exclusive to the validation study), and another ramped test (for validation only). After that, the gold-standard VO$_{2}$max data was collected. For BBVS participants, similar sensor and anthropometric data were collected in the same manner as in Fenland.

\subsection{Feature Extraction} \label{4.3}
All heart rate data collected during lab-visit underwent pre-processing for noise filtering~\citep{noise-filter}. If participants had fewer than 72 hours of concurrent wear data (three full days of recording) or inadequate individual calibration data, they were eliminated from the study (treadmill test-based data). Non-wear periods were excluded from the analyses through non-wear detection procedures. This pre-processing algorithm discovered lengthy durations of non-physiological heart rate and extended periods of no movement reported by the device's accelerometer (\textgreater 90 minutes). We used the calculation 1 MET = 71 J/min/kg (3.5 ml O2 min1 kg1) to convert movement intensities into standard metabolic equivalent units (METs). These conversions were then used to classify intensity levels, with behaviors less than 1.5METs classed as sedentary, those between 3 and 6 METs as moderate to vigorous physical activity (MVPA), and those greater than 6METs as vigorous physical activity (VPA). Since time can greatly impact physical activities, we encoded the sensor timestamps using cyclical temporal features \citep{Spathis_self_supervision}. 
Additionally, given the sensors' high sampling rate (1 sample/minute) after matching the HR and Acceleration modalities, learning patterns from such a lengthy sequence (a week's worth of sensor data contains more than 10,000 time steps) is unfeasible, even with the most powerful sequence models. Therefore, instead of segmenting the data into time windows for temporal data processing, we downsampled the wearable signals (Accelerometer and ECG) by a ratio of 15 to decrease the sequence length to 600 time steps. \imt{After that, we normalized the data by performing min-max scaling on all input features (sequence-wise for sensor data and column-wise for metadata).} Then, each feature vector with 19 features combining 
time series and metadata were put into various deep neural networks. 
A detailed view of these features is provided Table~\ref{tab:variables} in Appendix~\ref{apd:first}.

\section{Hyper-parameters}
\label{apd:first}

In this section, we provide the details of the feature choices of the input feature vectors \(\bm{X}\) and \(\bm{M}\). A detailed view of these features is provided in Table~\ref{tab:variables}. \imt{Further, we provide the details of the hyperparameters in Table ~\ref{table:hyper}.}

\begin{table}[H]
  \caption{\textbf{Description of the features/variables used in our analysis as inputs to the models.} The features with asterisks(\textbf{*}) are time-series and remaining are metadata. The final set of features is 19.}
  \scalebox{0.75}{
  \begin{tabular}{@{}llll@{}}
  \toprule
  \multicolumn{3}{l}{\textbf{Features/Variables}} & \textbf{Description} \\ 
  \midrule
  \multicolumn{3}{l}{\textbf{Sensors}} &  \\
  & \multicolumn{2}{l}{Acceleration*} & Acceleration measured in m\textit{g} \\
  & & & \\
  & \multicolumn{2}{l}{Heart rate (HR)*} & Mean HR resampled in 15sec intervals, measured in BPM \\
  & & & \\
  & \multicolumn{2}{l}{Heart Rate Variability (HRV)*} & \begin{tabular}[c]{@{}l@{}}HRV calculated by differencing \\ the second-shortest and the second-longest inter-beat interval \\ (as seen in \cite{Faurholt-Jepsen2017}), measured in ms\end{tabular}  \\
  & & & \\
  & \multicolumn{2}{l}{\begin{tabular}[c]{@{}l@{}}Acceleration-derived \\ Euclidean Norm Minus One (ENMO)*\end{tabular}} & \begin{tabular}[c]{@{}l@{}}ENMO-like variable (Acceleration/0.0060321) + 0.057 \\ 
 (as seen in  \cite{white2016estimation})\end{tabular} \\
  & & & \\
  & \multicolumn{2}{l}{\begin{tabular}[c]{@{}l@{}}Acceleration-derived \\ Metabolic Equivalents of Task (METs)*\end{tabular}} & \\
  & & Sedentary* & If Accelerometer \textless 1, take daily count and average \\
  & & Moderate to Vigorous* & If Accelerometer \textgreater{}= 1,  take daily count and average \\
  & & Vigorous* & If Accelerometer \textgreater{}= 4.15, take daily count and average \\
  \multicolumn{3}{l}{\textbf{Anthropometrics}} & \\
  & \multicolumn{2}{l}{Age} & Age, measured in years \\
  & \multicolumn{2}{l}{} & \\
  & Sex & & Sex is binary (female/male) \\
  & \multicolumn{2}{l}{} & \\
  & Weight & & Weight, measured in kilograms \\
  & \multicolumn{2}{l}{} & \\
  & Height & & Height, measured in meters.centimeters \\
  & \multicolumn{2}{l}{} & \\
  & Body Mass Index (BMI) & & BMI is calculated by Weight/$(\text{Height}^2)$, measured in kg/$m^2$ \\
  \multicolumn{3}{l}{\textbf{Resting Heart Rate}} & \\
  & \multicolumn{2}{l}{Wearable-derived RHR} & \begin{tabular}[c]{@{}l@{}}RHR is calculated by averaging the 4th, 5th, and 6th minute \\ of the baseline visit and adding to that the Sleeping Heart Rate\\ that has been inferred by the wearable device. \cite{gonzales2020resting} \end{tabular} \\
  \multicolumn{3}{l}{\textbf{Seasonality}} & \\
  & \multicolumn{2}{l}{Month of year} & \begin{tabular}[c]{@{}l@{}}The month number is used along with a coordinate encoding that \\ allows the models to make sense of their cyclical sequence.\end{tabular} \\ 
  \bottomrule
  \end{tabular}
  }
  \label{tab:variables}
\end{table}

\begin{table*}
  \centering 
  \small 
  \caption{\imt{\textbf{Hyper-parameter tuning}.
  }}
  \begin{tabular*}{\textwidth}{l @{\extracolsep{\fill}} llll}
  \toprule
      \textbf{Parameter}& \textbf{Search space} & \textbf{Selected value}  \\
    \midrule
    Dropout &\{0.2,0.3\} & 0.3 \\
    Optimizer & Adam &  Adam\\
    Learning rate &\{1e-2, 1e-3\} & 1e-3 \\
    Epochs & [0,100] & Early stopping\\
    Batch size & \{8,26,32\} & 8 \\
    $\alpha$ & \{0.01,0.02,0.03\} & 0.01 \\
    $\lambda$ &  \{(0.9, 0.1), (0.8, 0.2), (0.7, 0.3), (0.6, 0.4), (0.5, 0.5)\} & (0.9,0.1)\\
    \bottomrule
  \end{tabular*}

  \label{table:hyper} 
\end{table*}


\end{document}